# Structural Controllability and Observability in Influence Diagrams

**Brian Y. Chan**[*] and **Ross D. Shachter**[†]
Department of Engineering-Economic Systems
Stanford University
Stanford, CA 94305-4025

## Abstract

Influence diagram is a graphical representation of belief networks with uncertainty. This article studies the structural properties of a probabilistic model in an influence diagram. In particular, structural controllability theorems and structural observability theorems are developed and algorithms are formulated. Controllability and observability are fundamental concepts in dynamic systems (Luenberger 1979). Controllability corresponds to the ability to control a system while observability analyzes the inferability of its variables. Both properties can be determined by the ranks of the system matrices. Structural controllability and observability, on the other hand, analyze the property of a system with its structure only, without the specific knowledge of the values of its elements (Lin 1974, Shields and Pearson 1976). The structural analysis explores the connection between the structure of a model and the functional dependence among its elements. It is useful in comprehending problem and formulating solution by challenging the underlying intuitions and detecting inconsistency in a model. This type of qualitative reasoning can sometimes provide insight even when there is insufficient numerical information in a model.

## 1 Introduction

**Influence diagram** is a graphical representation for probabilistic and decision models. It was developed by Howard and Matheson (Howard 1984). Many approaches have been explored to analyze an influence diagram (Shachter 1988) since then.

[*]also with IBM Almaden Research Center, San Jose, CA 95120-6099. e-mail: bchan@almaden.ibm.com; phone: (408)927-2595

[†]e-mail: shachter@sumex-aim.stanford.edu; phone: (415)723-4525

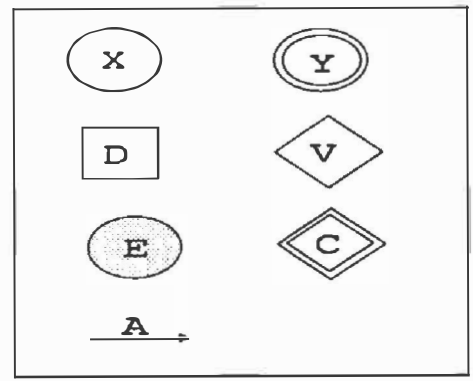

Figure 1: Components of an influence diagram

An influence diagram contains directed arcs and nodes which together represent probabilistic and deterministic variables, decisions, objectives and the functional relations among them as shown in figure 1:

- X = probabilistic variable
- Y = deterministic variable
- V = value node
- D = decision node
- A = arc
- E = observed evidence
- C = target node of control

Here is a simple model represented with an influence diagram: In a manufacturing process, a wafer needs to be set at a particular temperature before a chemical bath operation. The wafer is heated in an oven in advance, however, there is some heat loss during the transportation of the wafer from the oven to the chemical bath. The uncertainty lies in the fact that the heat loss is a probabilistic distribution, it varies with the room temperature and transporting time. The situation is modeled with an influence diagram as shown in figure 2.

Before any detail numerical analysis, it is prudent to verify that a model is robust. As indicated in the diagram, wafer temperature is a function of the oven temperature and heat loss. The wafer temperature cannot be set accurately without better understanding and control of the heat loss. It is fruitless trying to improve the process if this critical issue is ignored. The major effort should be focused on the maintenance of the room temperature and reduction of transportation time rather than the design of a more accurate oven. This type of information can be deduced from the structure (nodes and arcs) of the model without knowing its exact functions and numerical values. Such structural analysis is even more important in large influence diagrams consisting of thousands of arcs and nodes.

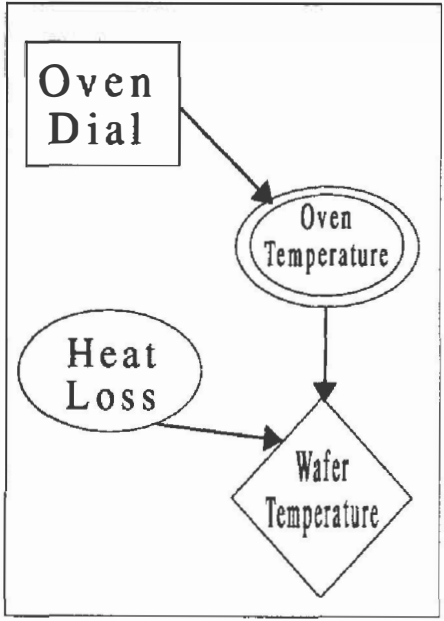

Figure 2: Wafer heating problem

Many linear and time-invariant **dynamic systems** can be conveniently expressed with state space equations (Luenberger 1979):

$$\vec{x}(t+1) = A(t)\vec{x}(t) + B(t)\vec{u}(t)$$

where $\vec{x}(t)$ and $\vec{u}(t)$ represent, respectively, the state and input variables at time $t$, whereas $A$ and $B$ are system matrices.

As an illustration, suppose there are 3 different products in a factory: X, Y, and Z. Each year, X can only be purchased. Y can be made from either X, Y, or Z. And Z can only be made from X.

$$\begin{bmatrix} X(t+1) \\ Y(t+1) \\ Z(t+1) \end{bmatrix} = \begin{bmatrix} 0 & 0 & 0 \\ \alpha_1 & \alpha_2 & \alpha_3 \\ \alpha_4 & 0 & 0 \end{bmatrix} \begin{bmatrix} X(t) \\ Y(t) \\ Z(t) \end{bmatrix} + \begin{bmatrix} 1 \\ 0 \\ 0 \end{bmatrix} U(t)$$

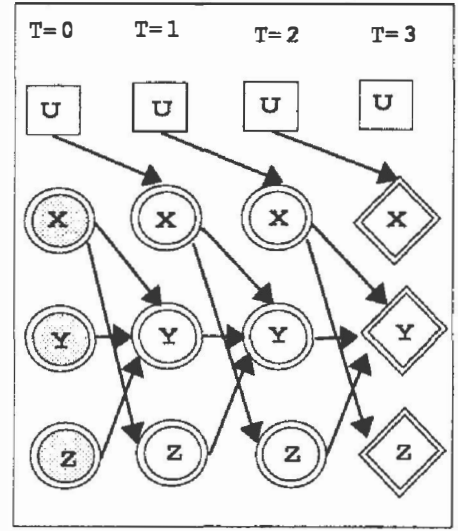

Figure 3: Products in a factory

If a dynamic graph (Murota 1987, Yamada 1990) of the above model is drawn based on the influence diagram notation, it will look like figure 3. As we will see later in the paper, this system is controllable.

Controllability and observability are fundamental concepts in dynamic systems. The $n$th order dynamic system $x(k+1) = Ax(k) + Bu(k)$ is **controllable** if for $x(0) = 0$ and any given $n$ vector $x_1$ there exists a finite index $N$ and a sequence of inputs $u(0), u(1), ..., u(N-1)$ such that this input sequence, applied to the system, yields $x(N) = x_1$ (Luenberger 1979). The system is **structurally controllable** if and only if $\forall \epsilon > 0$, there exists a completely controllable system $y = A_1 x + B_1 u$, of the same structure[1] as $y = Ax + Bu$ such that $\| A_1 - A \| < \epsilon$ and $\| B_1 - B \| < \epsilon$ (Lin 1974).

Consider the following example: $y = ax + bu$. $y$ is called structurally controllable because it can be set to any value by changing $u$, except in very rare coincidence such as $b = 0$.

Observability is a dual concept of controllability. The dynamic system $x(k+1) = Ax(k), y(k) = Cx(k)$ is completely **observable** if there is a finite index $N$ such that knowledge of the outputs $y(0), y(1), ..., y(N-1)$ is sufficient to determine the value of the initial state $x(0)$ (Luenberger 1979). The system is **structurally observable** if and only if $\forall \epsilon > 0$, there exists a completely observable system $x(k+1) = A_1 x(k)$, $y_1(k) = C_1 x(k)$ of the same structure as $x(k+1) = Ax(k), y(k) = Cx(k)$ such that $\| A_1 - A \| < \epsilon$ and $\| C_1 - C \| < \epsilon$.

---

[1] A dynamic system $y = Ax + Bu$ has the same structure as another system $y = A_1 x + B_1 u$, of the same dimensions, if for every fixed zero entry of the matrix $(A \mid B)$, the corresponding entry of the matrix $(A_1 \mid B_1)$ is fixed zero and vice versa.

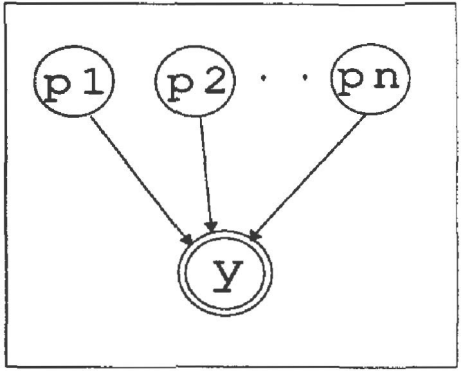

Figure 4: Deterministic function

Influence diagrams have non-stationary[2] and probabilistic variables that traditional dynamic systems do not.

## 2  Structural Observability in Influence Diagram

The relationship of a deterministic node $y$ with its parents $p_i (1 \leq i \leq n)$ can be modeled by the function $y = f(p_1, p_2, ..., p_n)$. (See figure 4)

And the relationship of a probabilistic node $x$ with its parents $p_i (1 \leq i \leq n)$ can be modeled by the function $x = f(p_1, p_2, ..., p_n, \vec{\epsilon})$, where $\vec{\epsilon}$ is an uncontrollable variable. (See figure 5)

**Definition 1:** A node is **known or observed** if its value has been determined from observation or deduction.

**Definition 2:** A node is **observable** if its value can be deduced from the information of other (observed) nodes.

**Definition 3:** Given that $x \in R^{n+m}$, $y \in R^n$, $m \geq 0$, $n \geq 1$, and a deterministic function $f : R^{n+m} \to R^n$, $f$ is **generically inferable** if $x$ can be uniquely determined from knowing the values of $y$ and any $m$ of $n$ elements of $x$, except in some rare coincidence. $x$ is said to be **structurally observable** given the value of $y$. An important special case is when $m = 0$, $f : R^n \to R^n$ is just the ordinary one to one function, and $x$ is invertible from $y$.

---

[2]Stationary relations are relations that stay the same in each time period. Linear system equations have stationary relations. Non-stationary relations are those that can be different in each time period.

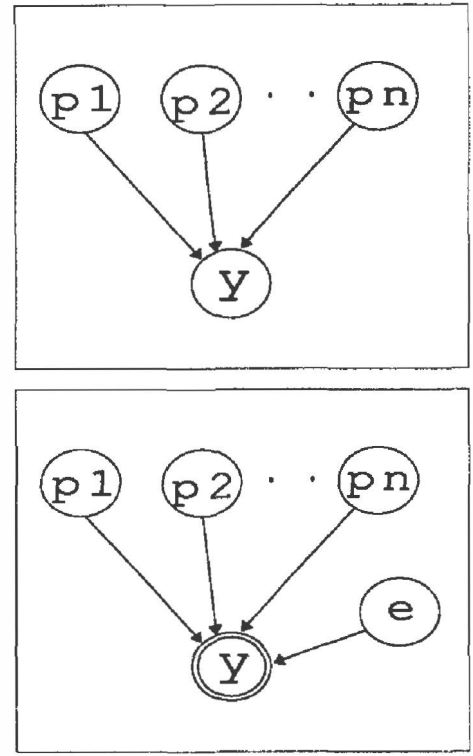

Figure 5: Representations of a Probabilistic Function

Here are some examples to illustrate the concept of generic inferability:

1. $y = x_1 + x_2$, is a generically inferable function, $x_1$(or $x_2$) can be determined if $y$ and $x_2$(or $x_1$) are known.

2.
$$\begin{bmatrix} y_1 \\ y_2 \\ y_3 \end{bmatrix} = \begin{bmatrix} \alpha_1 & 0 & \alpha_2 \\ \alpha_3 & \alpha_4 & \alpha_5 \\ \alpha_6 & \alpha_7 & 0 \end{bmatrix} \begin{bmatrix} x_1 \\ x_2 \\ x_3 \end{bmatrix}$$

$x_1, x_2$ and $x_3$ can be uniquely determined given $y_1, y_2, y_3$ are known, except in the rare coincidence that the determinant of the matrix is zero.

It is obvious that different functional classes such as linear, quadratic, Boolean, and so forth in a model will prescribe different system behavior. Since we are analyzing the structural properties of a model without the knowledge of its specific functions nor values, we need to tighten the functional domain slightly in order to derive any meaning results. We assume that the models we are studying have generically inferable functions. Most linear functions and many nonlinear functions have such property.

### 2.1  Structural observability theorems

**Theorem 1:** The value of a deterministic node are observable if the values of all its parents are known. (See figure 6)

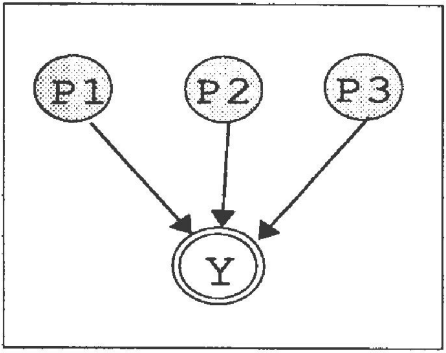

Figure 6: All parents known case

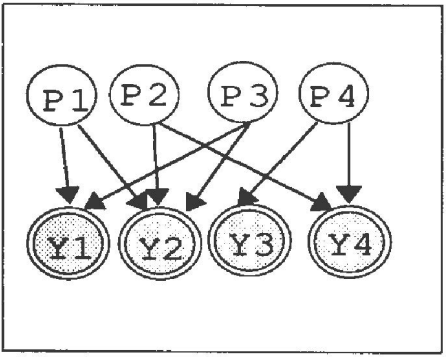

Figure 7: N by N case

**Proof:** A deterministic node $y$ is a function of its parents $p_i (1 \leq i \leq n)$, that is $y = f(p_1, p_2, ..., p_n)$.

**Theorem 2:** If the functions are generically inferable, a set $\mathcal{P}$ of k parent nodes are structurally observable given a set $\mathcal{S}$ of k deterministic children nodes (See figure 7) if

1. The values of all nodes in $\mathcal{S}$ are known.
2. All unknown parents of $\mathcal{S}$ are included in $\mathcal{P}$.
3. There exists a complete matching from $\mathcal{P}$ to $\mathcal{S}$.

**Proof:** Directly from definition of generic inferability in the m=0 case.

Consider a simple $f: R^1 \rightarrow R^1$ case: Let $x$ = result of a coin flip, there is a 50-50 chance of a head or tail. And let $y$ = winning based on the result of coin flip, receives one dollar if it is a head and nothing if it is a tail. (See figure 8)

$x$ is a probabilistic random variable and $y$ is a deterministic variable given the result of $x$. Now if the result of the flip is known, the winning can be determined. On the other hand, if you know whether any money is received, the result of the coin flip (which is a probabilistic random variable) can be inferred.

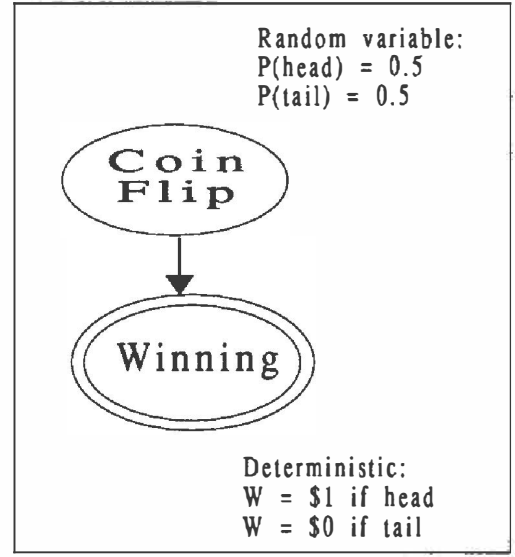

Figure 8: Inferring Data in Coin Flipping

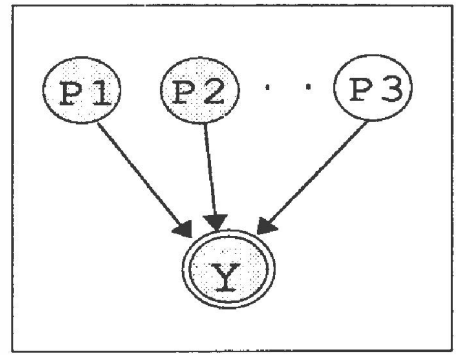

Figure 9: All but one case

**Corollary 1:** The value of a parent node $p$ is structural observable (See figure 9) if:

1. the value of its deterministic child $y$ is known, and
2. the values of all parents of $y$ except $p$ are known.

**Corollary 2:** If a chain is formed with several deterministic nodes, all nodes in the chain are observable if any one of them is known.

### 2.2 Algorithm to determine structural observability

1. For each unknown deterministic node, check if all its parents are known, if they are, mark the child node as 'observable' and mark the arcs between parents and child as 'blocked'. A 'blocked' arc means this relation cannot provide further observability information in the influence diagram.

2. For all known deterministic nodes:
   (a) Partition the deterministic nodes into family equivalence[3] classes.
       - If two deterministic nodes are siblings[4] then they must be in the same family class.
   (b) For each family class:
       i. Form a bipartite graph with all the known deterministic nodes and their unknown parents.
       ii. If there is a complete matching of k parents with k deterministic known children, and all unknown parents of these k children are included in the matching, then these k parents can be marked as 'observable' and the associated arcs marked as 'blocked'.
       
       If k parents are covered by more than k deterministic children, this situation corresponds to the generating of several deterministic children from the same parent. Redundant data should be verified to check system consistency.
       iii. The search is repeated until no more such set of k parents can be found.
   (c) If there is any addition to the total number of observable nodes, then go to step one again. Otherwise stop, all the observable nodes have been determined.

### 2.2.1 Observability Example

In figure 10, given that D, F, G and H are observed, all other nodes except J are observable.

| Observed | Inferable | Reason |
|---|---|---|
| D, F, H | E | corollary 1 |
| E | A | corollary 1 |
| F, G | B, C | theorem 2 |
| E, F, G | I | theorem 1 |

### 2.2.2 Option Investment Example

The price of a stock option is a function of the current stock price, the strike price, time to expiration, risk free interest rate, and stock volatility[5]. Among these five factors, the first four can be measured objectively from market data. The last one, stock volatility, is subjective and is based on people's beliefs of the fluctuation of the stock price in the future.

One option strategy that traders frequently play is based on the volatility discrepancy between people's beliefs and the historical data. Many traders think that the implied volatility (which is based on people's beliefs) should be within a narrow range of the historical volatility data. If the two differ beyond a certain margin, there is an investment opportunity.

---
[3]Equivalence relation is transitive, symmetric and reflexive.
[4]Two nodes are siblings if they have one or more parents in common.
[5]Black-Scholes formula, see any investment reference books such as (Hull 1989) for detail.

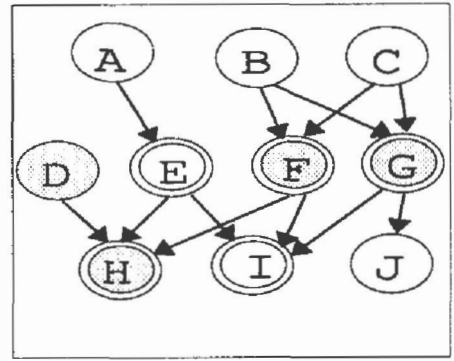

Figure 10: Observability example

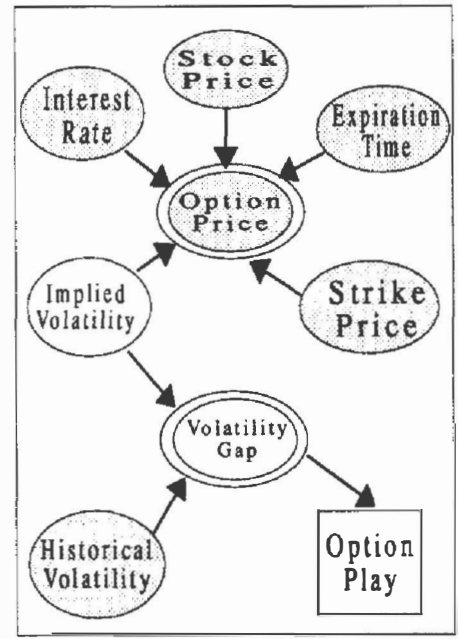

Figure 11: Option Investment Decision

The investment decision is modeled with an influence diagram as seen in figure 11. Since the option's price is listed, and all the factors except the volatility are known, the implied volatility is observable (from corollary 1: all but one case). The historical data can be obtained from market database, therefore the volatility gap is determined (from theorem 1: all parents known case). A decision can now be made whether to play the strategy or not based on the input from the volatility gap.

# 3 Structural Controllability in Influence Diagram

**Definition 4:** A node is **controllable** if its value can be set to any value either directly or by changing the values of some other nodes.

**Definition 5:** Given that $x \in R^{n+m}$, $y \in R^n$, $m \geq 0$, $n \geq 1$, and a deterministic function $f : R^{n+m} \to R^n$, $f$ is **generically nimble** if $y$ can be set to any value from

- knowing the values of any $m$ of $n+m$ elements of $x$, and
- adjusting the remaining $n$ elements of $x$

except in some rare coincidence cases. $y$ is said to be **structurally controllable** by $x$.

An important special case is when $m = 0$, the function $f$ then is just the onto functions.

## 3.1 Structural controllability theorems

**Proposition 1:** A decision node is controllable.

**Theorem 3:** Assume the functions are generically nimble, a node $v$ is controllable (See figure 12) if:

1. $v$ is a value node or a deterministic node.
2. $v$ is reachable[6] from a decision node $d$.
3. All the nodes $x_i (i \geq 0)$ on the directed path from $d$ to $v$ are deterministic nodes.
4. Let $X$ be the set that contains all $x_i (i \geq 0)$ on the directed path and $P$ be the set that contains all parents of $x_i \in X$. Then all nodes in $P - X$ have to be structurally observable or controllable by decision nodes other than $d$.

**Proof:** $v$ is function of $x$ and $x$ is a function of $d$. The generically nimble property is transitive[7]. Therefore, if the other parameters are observed, $v$ can be set to any value by changing $d$.

**Theorem 4:** If the functions are generically nimble, a set $\mathcal{V}$ of nodes $v_i (1 \leq i \leq n)$ are structurally controllable if

1. $\mathcal{V}$ does not contain probabilistic nodes.
2. There are at least $m(m \geq n)$ distinct decision nodes.
3. A set of $n$ deterministic node disjoint paths[8] from decision nodes $d_i (1 \leq i \leq m)$ to $v_i (1 \leq i \leq n)$ can be found.

---

[6]A node $v$ is reachable from a node $d$ if there is a directed path starting at $d$ which contains $v$ (Shachter 1990).
[7]for detail proof, please see (Chan 1992).
[8]Node disjoint paths are directed paths that do not visit the same node.

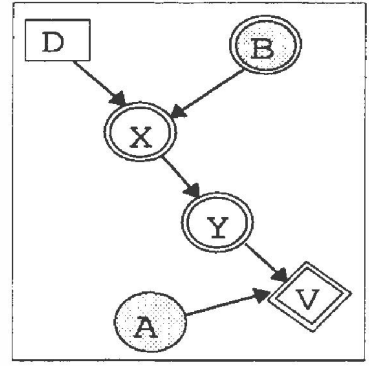

Figure 12: Controllable value node V

4. Let $Y$ be the set that contains all nodes on the above node disjoint paths, and $P$ be the parents of the nodes in $Y$. Then all nodes in $P - Y$ (in $P$ but not in $Y$) have to be structurally observable or themselves controllable with decision nodes other than $d_i (1 \leq i \leq m)$.

Proof: Please see (Chan 1992).

## 3.2 Algorithm to determine structural controllability

1. Check if the target set of control contains value nodes and/or deterministic nodes only. If not, the set, as a whole is not structurally controllable.

2. Check that the number of decision nodes is greater than or equal to the number of nodes in the target set.

3. Check that decision nodes are not predecessors of any observed nodes. If they are, the values of those decision nodes may be observable but the nodes cannot be used as control, since decisions have already been predetermined.

4. Construct node disjoint paths with max-flow method. Decision nodes are sources, and the nodes in the target set are the sinks. The flow capacity of every deterministic node is one and the flow capacity of every probabilistic node is zero. Target set is not structurally controllable if not enough node disjoint paths can be found.

5. Let $Y$ contains all the nodes on the node disjoint paths, and $P$ contains all parents of nodes in $Y$. Check that all nodes in $P - Y$ are structurally observable (with the structural observability algorithm described in the previous section), or controllable by some other additional decision nodes.

6. If all of the above conditions are satisfied, then target set of control is structurally controllable. If not, go to step 4 and try to find other node disjoint paths again.

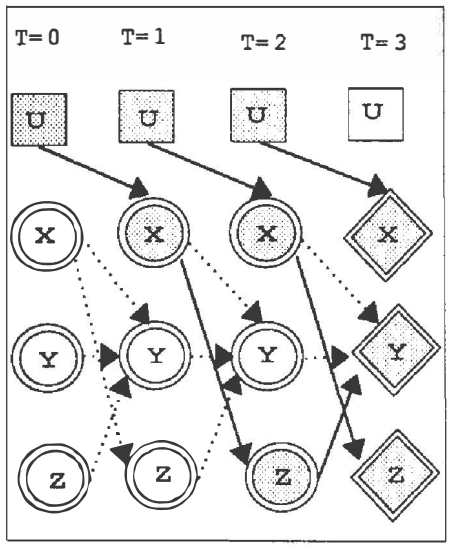

Figure 13: Products in a factory example

### 3.2.1 Products in a factory example

Consider the products in a factory dynamic system discussed earlier in section 1. Since the initial states of X, Y, and Z (at time 0) are given, all the values of X, Y, and Z at subsequent times are observable. Furthermore, we can find three node disjoint paths from the controls to the values:

U(2) to X(3)
U(1) to X(2) to Z(3)
U(0) to X(1) to Z(2) to Y(3)

Therefore, the system is structural controllable[9] At time T=3, X, Y, and Z can be set to any values we want (See figure 13).

### 3.2.2 Wafer heat loss example

Let us consider the wafer heat loss problem again. As before, the heat loss is a function of the room temperature and transportation time. But now the transportation is automated with a conveyer belt and therefore the transporting time is constant. Furthermore, the room temperature is relative stable and can be measured. The modified model is drawn in figure 14.

Since the room temperature and transportation time are observed, the heat loss can be calculated. In addition, there is a deterministic node path from the oven dial to the wafer temperature. The oven temperature can be raised slightly higher to compensate exactly

---

[9]For partial structural controllability problems in dynamic systems, an additional step to verify global nimble property might be needed (Chan 1992). Dynamic systems have stationary functions that replicate in each period, they lend themselves more easily into mutual dependence cases (Murota 1990).

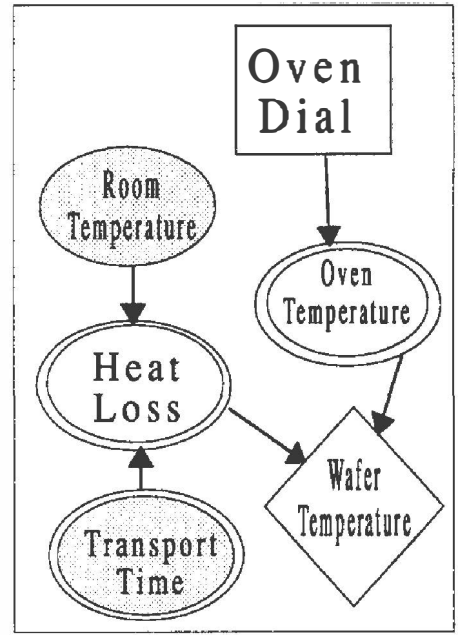

Figure 14: Wafer heating problem

for the heat loss. Therefore the wafer temperature is controllable.

## 4 Conclusions

We have described the structural controllability and observability theorems in influence diagrams. The ability to analyze a probabilistic model with its structure is important in the design and comprehension of a system. It is especially useful in model validation and rapid prototype constructions in large probabilistic systems.

### Acknowledgements

We benefitted greatly from the comments and suggestions of Michael Fehling, Gene Franklin, David Luenberger, Edison Tse and the two anonymous referees.